\documentclass{article}
\usepackage{ijcai23}

\usepackage{times}
\usepackage{soul}
\usepackage{url}
\usepackage[hidelinks]{hyperref}
\usepackage[utf8]{inputenc}
\usepackage[small]{caption}
\usepackage{graphicx}
\usepackage{amsmath}
\usepackage{amsthm}
\usepackage{booktabs}
\usepackage{algorithm}
\usepackage{algorithmic}
\usepackage[switch]{lineno}

\usepackage{natbib}
\usepackage{subcaption}
\usepackage{microtype}
\usepackage{multirow}
\usepackage{amsfonts}
\usepackage{pifont}
\newcommand{\cmark}{\ding{51}}
\newcommand{\xmark}{\ding{55}}

\title{Depth-Relative Self Attention for Monocular Depth Estimation}

\author{
Kyuhong Shim\and
Jiyoung Kim\and
Gusang Lee\And
Byonghyo Shim
\affiliations
Department of Electrical and Computer Engineering, Seoul National University, Korea
\emails
\{khshim, jykim, gslee, bshim\}@islab.snu.ac.kr
}

\begin{document}
\maketitle
\begin{abstract}
Monocular depth estimation is very challenging because clues to the exact depth are incomplete in a single RGB image.
To overcome the limitation, deep neural networks rely on various visual hints such as size, shade, and texture extracted from RGB information.
However, we observe that if such hints are overly exploited, the network can be biased on RGB information without considering the comprehensive view.
We propose a novel depth estimation model named RElative Depth Transformer (RED-T) that uses relative depth as guidance in self-attention.
Specifically, the model assigns high attention weights to pixels of close depth and low attention weights to pixels of distant depth.
As a result, the features of similar depth can become more likely to each other and thus less prone to misused visual hints.
We show that the proposed model achieves competitive results in monocular depth estimation benchmarks and is less biased to RGB information.
In addition, we propose a novel monocular depth estimation benchmark that limits the observable depth range during training in order to evaluate the robustness of the model for unseen depths.

\end{abstract}
\section{Introduction}\label{sec:intro}

Depth estimation, a task to estimate the distance from the viewpoint, is one of the most important tasks in computer vision having a variety of applications such as autonomous driving~\citep{auto_drive_a,auto_drive_b}, object localization~\citep{localization_b}, 3D reconstruction~\citep{3d_recon_b}, to name just a few.
Due to the cost and power consumption of depth measuring sensors (e.g., LiDAR, Time-of-Flight), a single RGB image has been used for this task in many real-world applications~\citep{real-time-mde,mobile-mde,fast-depth}.
The major difficulty of this task, \textit{monocular depth estimation} (MDE), is that the task is ill-posed since there are multiple answers for the given scene.
Recently, deep neural networks alleviated this problem by exploiting diverse visual clues such as relative size, brightness, patterns, and vanishing point extracted from an RGB image.
It has been shown that these visual clues, collectively called ``\textit{visual hint}s'', are useful in predicting the depth~\citep{depth_cues,depth_review}.

\begin{figure*}[t]
    \centering
    \includegraphics[width=0.88\textwidth]{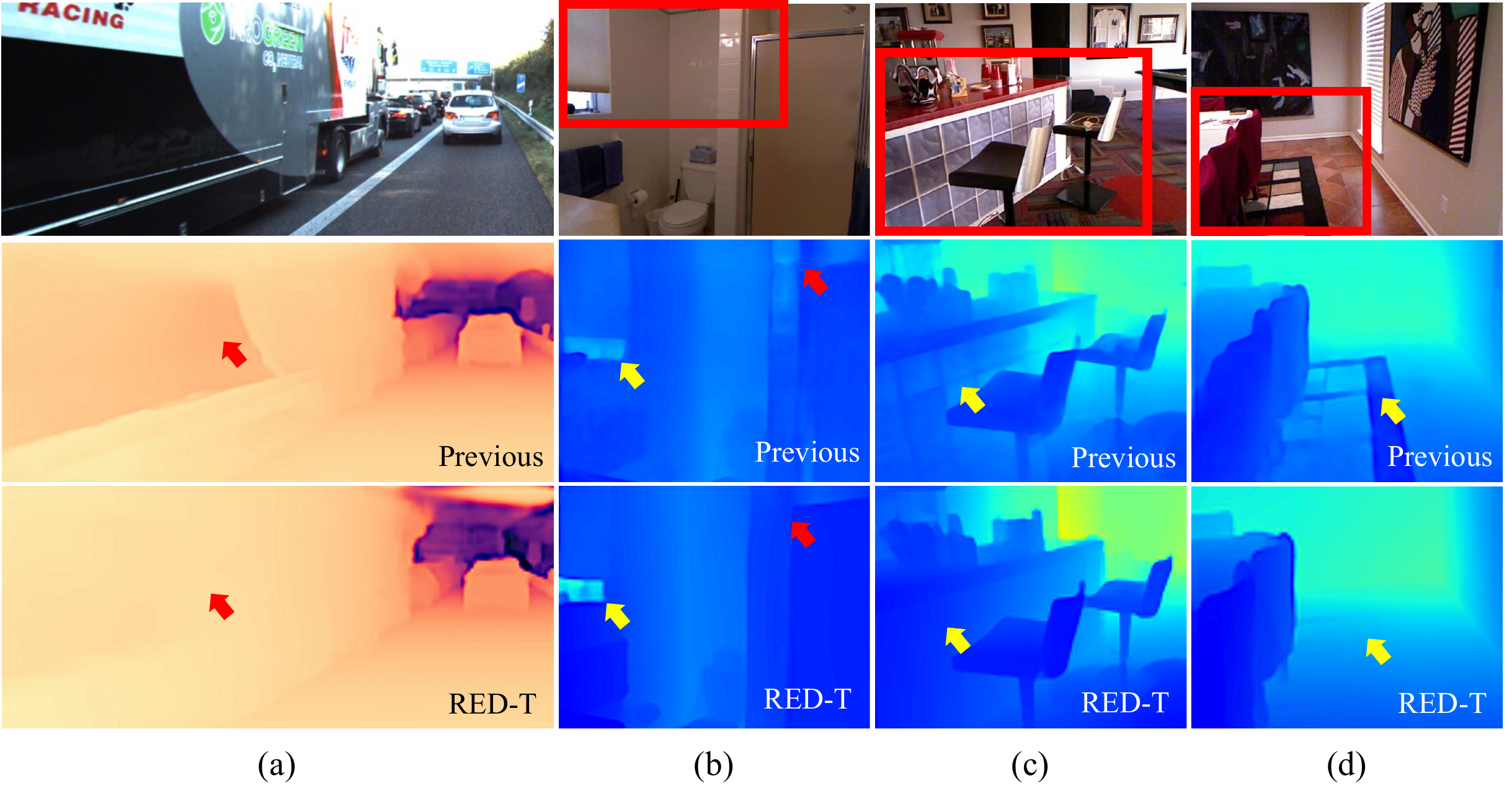}
    \caption{
    Examples of the \textbf{misused visual hints}, referred to as \textbf{visual pits}.
    Visual pits can disturb the correct depth estimation: (a) dark paint on the truck's surface, (b) sunlight and its reflection on the wall, (c) square pattern of the kitchen counter, and (d) colorful pattern of the carpet.
    As observed in the second row, the previous depth estimation model~\citep{adabins} suffers from undesirable flaws caused by visual pits.
    In contrast, the proposed model is robust to such visual pits.
    The results are best viewed in PDF.}
    \label{fig:comparison}
\end{figure*}

In order to improve the quality of visual hints, a pre-trained network referred to as `backbone' has been widely used.
Recently, depth estimation performance has been substantially improved~\citep{depthformer,newcrf} due to the diverse and complex RGB-based visual features obtained from the large-scale backbone networks~\citep{rich_feature,bit,swin}.
However, we observe that some visual information such as painted surfaces, a patterned carpet, and reflected sunlight sometimes provide false signals to the network and degrade the accuracy of the predicted depth (see Figure~\ref{fig:comparison}).
While the visual hints are useful to some extent, they would do more harm than good when models become overly dependent on such information.
In the sequel, we call the visual hints that confuse the model and therefore have an adverse effect on the depth estimation as ``\textit{visual pit}s''.
For example, in Figure~\ref{fig:comparison}(a), the dark paint of the truck affects the model such that the truck appears farther away than it really is.
Clearly, visual pits can be a potential risk factor for the autonomous driving system.

To reduce the negative effect of visual pits, we should design the system such that the extracted features are more related to depth while less dependent on RGB-based information.
In other words, we expect that the features corresponding to pixels of similar depths to be similar.
As an enabler to achieve this goal, we exploit the \textit{relative depth}, a difference between the depth of two pixels.
If the relative depth between two pixels is small, the model should generate similar features regardless of their RGB attributes and spatial distances in 2D image\footnote{In order to make a clear distinction between the distance in the 2D image and the real world, we exclusively use the terms `near/far' for the former, and `close/distant' for the latter.}.
When this property is satisfied, even though the two complementary information (i.e., RGB and depth) make contradictory predictions, we can use the relative depth as guidance in estimating the correct depth.
For example, in Figure~\ref{fig:comparison}(b), the model is confused by the reflected sunlight, resulting in an incorrect prediction that the upper part of the pillar is farther than its real depth.
Even in this case, using the relative depth between the lower and upper parts of the pillar is small, the model can figure out that both parts are actually at the same depth.

In this paper, we propose a novel MDE model referred to as \textbf{RElative Depth Transformer~(RED-T)}.
The key idea of RED-T is to exploit the relative depth as guidance in computing the self-attention weights.
To this end, we design the \textit{depth-relative attention} module on top of the backbone.
Using the relative depth information, this module modifies the self-attention weight in two steps.
First, we gather the relative depth information to determine which pixels should be similar in the feature domain.
Second, we adjust the self-attention weight based on the relative depth; large (small) attention weights for pixels of small (large) relative depths.
We expect that features with large attention weights are more or less similar to each other since the self-attention mechanism is basically a weighted sum of features.
In fact, through the proposed depth-aware self-attention process, the features corresponding to pixels with small relative depths will be close to each other in the feature domain, even if their RGB values are different.
Whereas, if the relative depth is large, the features would be distinct although their corresponding RGB attributes may look alike.

To demonstrate the negative effect of visual pits, we propose a new practical MDE environment termed \textbf{range-restricted MDE}.
In the conventional depth estimation benchmarks, the target depth range is the same for both training and evaluation.
To make things worse, the annotated depth data has limitations in range (e.g., 10\textit{m} for NYU-v2~\citep{nyu} and 80\textit{m} for KITTI~\citep{kitti} datasets).
However, in real-world scenarios, we should estimate the depth of distant objects correctly even if their depths are not specified in the training data.
When the model only learns the correlation between RGB attributes and limited depths, the model would inaccurately estimate the unseen depths to seen depths highly dependent on the RGB information, which will intensify the adverse influence of visual pits on such unseen depth range.
In the proposed environments, we erase the depth labels of a certain range as if they are not annotated in training data.
For example, for the data with a depth range of $0\sim 80$\textit{m}, we remove the training labels in $40\sim 80$\textit{m} and evaluate the model with the full range ($0\sim 80$\textit{m}).
In our evaluations, we show that RED-T predicts not only the learned depth but also the \textit{out-of-range} depth more accurately than previous state-of-the-art MDE models.

The contributions of this work are as follows:
\begin{itemize}
    \item We employ relative depth, a difference between the depth of pixels, as guidance to solve the problem of \textit{visual pit}.
    To the best of our knowledge, we are the first to tackle the negative effect of visual information in MDE.
    \item We propose a novel depth-relative attention that adjusts the self-attention weight based on the relative depth.
    In essence, the proposed mechanism guides feature such that the depth is more considered than RGB information.
    \item Using two MDE datasets (KITTI \& NYU-v2), we evaluate RED-T and show that the proposed RED-T outperforms the recent MDE models that use the same backbone in all metrics for the extremely competitive KITTI dataset.
    \item To evaluate the performance of MDE models in practical environments, we suggest new depth estimation scenarios that restrict observable depth range during training.
    We show that the depth-relative attention bias makes the model more robust in estimating unseen depth ranges.
\end{itemize}

\begin{figure*}[t]
    \centering
    \includegraphics[width=0.90\linewidth]{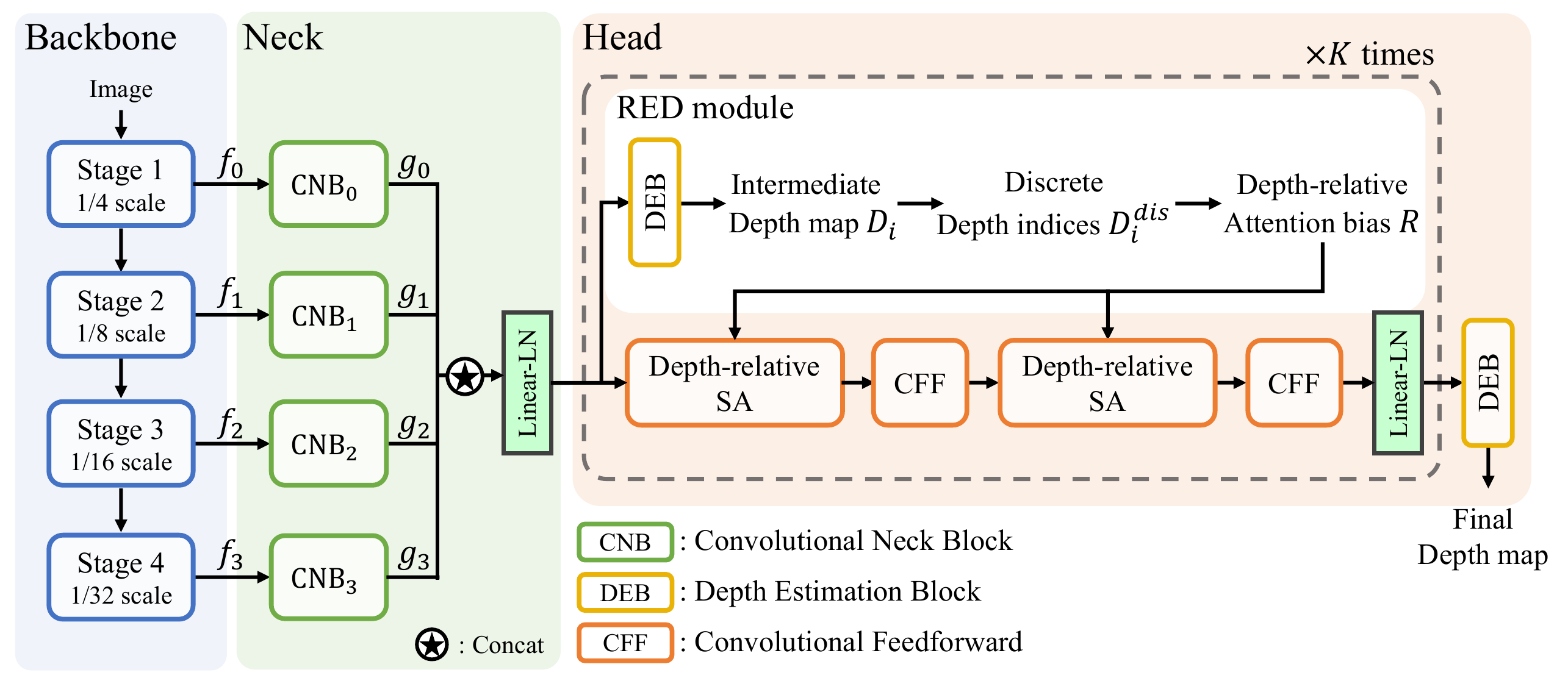}
    \caption{Overview of the proposed RElative Depth Transformer (RED-T).
    First, the backbone~(blue) extracts multi-scale features from the input RGB image.
    Second, the neck~(green) aggregates different scales at once.
    Finally, the head~(orange) iteratively refines the feature using Transformer blocks and generates the final depth map.
    Please see Appendix for the detailed structure of CNB, DEB, and CFF blocks.
    }
    \label{fig:overview}
\end{figure*}
\section{Related Work}\label{sec:related}

\paragraph{Adversarial Effect of Visual Pits}
Visual information extracted from an RGB image such as color, texture, style, and brightness is known to be helpful in object detection, semantic segmentation, and super resolution~\citep{detect_1,swinir,local}, etc.
However, this perceptual information is not always reliable, especially when the goal of the task is to generate output in a non-RGB domain with RGB input.
For example, in the image segmentation task, the pixels corresponding to the same class must produce the same mask even if their RGB values are different.
Previous work pointed out that the physical effects of illumination, shadow, shading, and highlights can cause considerable noise in the image segmentation output~\citep{seg_1}.
Likewise, in MDE, visual pits such as patterned surfaces, dark screens, and reflections in the mirror can disturb the depth estimation process.
To avoid the failures caused by visual pits, we exploit the relative depth information such that the MDE model can focus more on depth-related information.

\paragraph{Relative Depth}
Several studies have used relative depth information for depth estimation~\citep{rel_depth_related_1,rel_depth_related_2}, but their works are very distinct from ours.
The main difference to the referred papers is that 1) we investigated the ‘visual pit’ problem which has not been studied before and 2) we exploited the novel depth-relative attention bias instead of the position-relative bias.
We note that studies~\citep{rel_depth_related_1,rel_depth_related_2} that introduced relative depth in the feature extraction are difficult to apply when dense depth labels are not given.
For example, the former does not show the performance on KITTI, a dataset where only a few pixels are labeled sparsely.
Also, the latter mentioned that the training process was less reliable on KITTI as their labels are not annotated.
In contrast, as can be observed in restricted label experiments, our method works well on much sparser scenarios than the original KITTI.

\section{RED-T: RElative Depth Transformer}\label{sec:method}

In this section, we discuss three components, backbone, neck, and head, of the proposed RED-T.
Figure~\ref{fig:overview} illustrates the overall architecture of the model.

\begin{figure*}[t]
    \centering
    \includegraphics[width=0.90\linewidth]{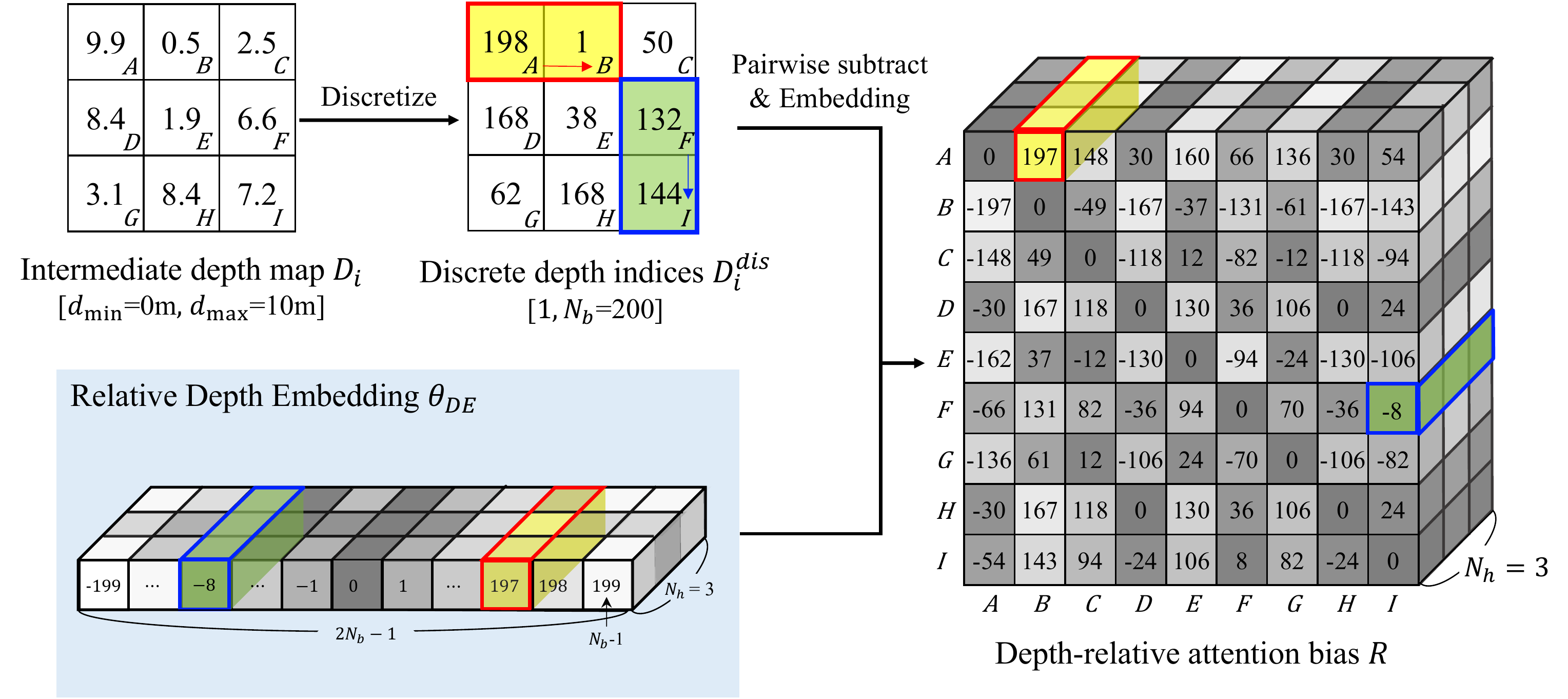}
    \caption{Computation of depth-relative attention bias $R$.
    The bias is added to self-attention weight to adjust the weight based on the relative depth between pixels.
    \textit{A, B, ... I} indicates the spatial location of each pixel.
    Here, we set $N_b=200$ and $N_h=3$ for better understanding.
    }
    \label{fig:rel_depth_bias}
\end{figure*}

\subsection{Monocular Depth Estimation}\label{ssec:mde}

Let $H$ and $W$ be the height and width of the image, then the MDE model takes a single RGB image $I \in \mathbb{R}^{H\times W\times 3}$ as input and returns the estimated depth map $D \in \mathbb{R}^{H \times W \times 1}$.
Each element of $D$ represents a distance $d$ from the viewpoint.
Because the ground truth depth map $D^{*}$ contains only a few annotated pixels, the loss is computed on those pixels in the training stage.

\subsection{Backbone: Position-relative Transformer}\label{ssec:backbone}

As a backbone, we use Swin Transformer (Swin)~\citep{swin}, a multi-stage Transformer whose self-attention (SA) is computed with non-overlapping local windows.
In Swin, SA between $n$ pixels is calculated as:
\begin{align}
               A_h ( Q_h, K_h, B_h ) &= \text{Softmax} \Big( \frac{Q_{h} K_{h}^{T}}{\sqrt{d_h}} + B_h \Big) \label{eq:rel_pos} \\
    \text{SA}_h (Q_h, K_h, V_h, B_h) &= A_h (Q_h, K_h, B_h) V_h
\end{align}
where $h$ is the attention head index over the total number of heads $N_h$ and $d_h$ is the attention head dimension.
$Q_h,K_h,V_h \in \mathbb{R}^{n\times d_h}$ are the query, key, value matrices for the $h\text{-th}$ attention head, respectively, and $A_h \in \mathbb{R}^{n \times n}$ is the attention weight.
To promote the spatial relationship between pixels, Swin adds the relative positional attention bias $B_h \in \mathbb{R}^{n \times n}$ to the attention weight.
Note that $B_h$ is unrelated to the content and depends only on the difference in \textit{coordinates} (i.e., spatial location) between pixels.

\subsection{Neck: Parallel Multi-scale Aggregation}\label{ssec:neck}

The neck performs parallel processing of the multi-scale backbone features with different scales and then stacks them together at the highest resolution (largest scale).
To ensure that the scale of features is the same, features are up-sampled to the highest resolution.
Let $f_{\{0,1,2,3\}}$ be the image features extracted from the backbone corresponding to $1/4, 1/8, 1/16, 1/32$ scales, then each image feature $f_i$ is passed through a convolutional block.
The block takes $i\text{-th}$ feature $f_i$ as an input and then returns the processed feature $g_i$.
The generated features $g_{\{0,1,2,3\}}$ are concatenated and passed through an additional linear layer followed by layer normalization.

Traditional feature pyramid network (FPN) merges multi-scale features one after another, from the smallest scale features to the largest scale ones~\citep{od_fpn,od_efficientdet,od_yolov3}.
Since FPN merges the multi-scale features sequentially, global information from small-scale features can be blurred during the hierarchical process~\citep{GCPANet,CPNet}.
This might cause a loss of the global information presumably obtained from low-resolution features in local pixels.
Our neck architecture overcomes the potential weakness by combining all scales simultaneously.

\subsection{Head: Depth-relative Transformer}\label{ssec:head}

The relative depth $r$ between two pixels $x_1$ and $x_2$ is the difference between their depth values $d_1$ and $d_2$, that is $r(x_1, x_2) = d_1 - d_2$.
To obtain the relative depth between every pair of pixels, each pixel should have its own depth value; however, such a dense depth map is not available during the training and even the GT map does not contain depth values for all pixels.
To deal with the issue, we generate the intermediate dense depth map prediction and use it to compute the relative depth information.
The relative depth information is then used to predict the enhanced depth map.
This process can be interpreted as self-guided bootstrapping; RED-T repeats this cycle multiple times ($K$ times) to improve the intermediate depth maps progressively.

\noindent The detailed process of each cycle is as follows:

\paragraph{Discretization}
In the $i\text{-th}$ iteration, the model produces an intermediate depth map $D_i$.
Since $D_i$ is a real-valued dense depth map, every pixel of $D_i$ has its own estimated depth value and thus every relative depth can be computed.
Then, we discretize depth values by uniformly splitting the min-max depth range, where the number of bins $N_{b}$ is a hyperparameter.
This discretization converts depth map $D_i$ into $D_i^{dis}$, as illustrated in Figure~\ref{fig:rel_depth_bias}.
Note that the number of possible relative depths after the discretization is $2N_{b} - 1$, from $- N_{b} +1 $ to $N_{b} - 1$. 
If we increase $N_b$ in the discretization process, a more fine-grained granularity of relative depth can be obtained.
We empirically observed that 128 bins are sufficient.

\paragraph{Parameterization}
We parameterize the possible relative depths as embedding parameters $\theta_{\text{DE}} \in \mathbb{R}^{(2N_{b}-1)\times N_{h}}$ (see Figure~\ref{fig:rel_depth_bias}).
The goal of this parameterization is to map a raw relative depth to a trainable parameter that can be simultaneously trained with other parameters.
By doing so, the effect of relative depth on the attention weight can be automatically adopted for performance during training.
Note that the parameter size of $\theta_{\text{DE}}$ is quite small (about 2K per each self-attention module) although different attention head uses different embedding parameters.

\paragraph{Pairwise subtraction \& Embedding}
For every pixel pair, we perform a pairwise subtraction of two discretized depth values from $D_i^{dis}$ and then take the corresponding embedding parameter from $\theta_{\text{DE}}$ to construct the depth-relative attention bias $R$.
For example, in Figure~\ref{fig:rel_depth_bias}, the pairwise subtraction outputs 197 by subtracting 198(A) and 1(B), which are discretized depths. Then, the vector corresponding to the index 197 is taken from $\theta_{\text{DE}}$ to (A, B) point of $R$. 
The $R$ represents the relationship between pixels in terms of their depth difference, or relative depth.
Note that each entry of $R$ only depends on the relative depth between pixels and not on their visual features.

\paragraph{Depth-relative Self-attention}
Instead of using the conventional \textit{relative positional} attention bias $B$ (see Eq.~\eqref{eq:rel_pos}), we incorporate the \textit{relative depth} attention bias $R$ as below:
\begin{equation}
    A_h (Q_h, K_h, R_h) = \text{Softmax} \Big( \frac{Q_{h} K_{h}^{T}}{\sqrt{d_h}} + R_h \Big).    \label{eq:rel_depth}
\end{equation}
This novel depth-relative SA mechanism encourages pixels of similar depth to focus more on each other.
By assigning higher attention weight to features of similar depth (small relative depth), the features can be more correlated to depth.
Thus, the model can less affected by visual pits such as patterns or colors.

\subsection{Other Details}\label{ssec:model_detail}

\paragraph{Relative Depth Computation}
From earlier trials, we find that the uniform separation of depth range works well and performs better than the log-uniform partitioning as suggested in DORN~\citep{dorn}.
Different depth-relative SA blocks equip their own relative depth embedding parameters $\theta_{\text{DE}}$ so that each SA can learn diverse depth relationships.

\paragraph{Training loss}
The total loss $L_{\text{total}}$ is the scale-invariant loss~\citep{Eigen} averaged over all intermediate depth maps $D_{i\in\{0, 1, ... K-1\}}$ and the final depth map $D_{K}$.
\vspace{0.2cm}
\begin{equation}
    L_{i} = \alpha \sqrt{ \frac{1}{T}\sum_{x=0}^{T-1}{h_{i,x}^2} - \frac{\lambda}{T}\big( \sum_{x=0}^{T-1}{h_{i,x}} \big)^2  }, \quad
    L_{\text{total}} = \sum_{i=0}^{K}\frac{L_{i}}{K+1} \nonumber
\end{equation}
where $h_{i,x}=\log {d_x^*} - \log {d_{i,x}}$ and $T$ is the number of valid GT labels.
The loss for each depth map is calculated by the same equation above, reducing the possibility of the wrong prediction being amplified through iterations.
We set $\lambda$=0.85 and $\alpha$=10 following previous works~\citep{adabins,newcrf}.

\section{Experiment}\label{sec:experiment}

\setlength{\tabcolsep}{12pt}
\begin{table*}[t]
    \centering
    \resizebox{1.0\linewidth}{!}{
    \begin{tabular}{c|c|cccccc}
        \toprule
        Method & Backbone & Abs Rel$\downarrow$ & RMSE$\downarrow$ & $\log{10}$ $\downarrow$ & $\delta_{1}\uparrow$ & $\delta_{2}\uparrow$ & $\delta_{3}\uparrow$ \\
        \midrule
        DORN~\citep{dorn} & ResNet-101 & 0.115 & 0.509 & 0.051 & 0.828 & 0.965 & 0.992 \\
        BTS~\citep{bts} & DenseNet-161 & 0.110 & 0.392 & 0.047 & 0.885 & 0.978 & 0.994 \\
        TransDepth~\citep{TransDepth} & R-50+ViT-B$\dagger$ & 0.106 & 0.365 & 0.045 & 0.900 & 0.983 & 0.996 \\
        DPT~\citep{vitdepth} & R-50+ViT-B$\ddagger$ & 0.110 & 0.357 & 0.045 & 0.904 & 0.988 & \textbf{0.998} \\
        Adabins~\citep{adabins} & E-B5+mini-ViT & 0.103 & 0.364 & 0.044 & 0.903 & 0.984 & \underline{0.997} \\
        \midrule
        NeWCRFs~\citep{newcrf} & Swin-L$\dagger$ & 0.095 & 0.334 & 0.041 & 0.922 & \textbf{0.992} & \textbf{0.998} \\ 
        DepthFormer~\citep{depthformer} & R-50-C$_{1}$+Swin-L$\dagger$ & 0.096 & 0.339 & 0.041 & 0.921 & 0.989 & \textbf{0.998}  \\ 
        BinsFormer$^{*}$~\citep{binsformer} & Swin-L$\dagger$ & \underline{0.094} & \underline{0.330} & \underline{0.040} & \underline{0.925} & 0.989 & \underline{0.997}  \\
        \midrule
        \textbf{RED-T (Ours)} & Swin-L$\dagger$& \textbf{0.091} & \textbf{0.328} & \textbf{0.039} & \textbf{0.926} & \underline{0.990} & \textbf{0.998} \\
        \bottomrule
    \end{tabular}}
    \caption{Depth estimation performance on \textbf{NYU-v2} dataset.
    The best and second results are in \textbf{bold} and \underline{underlined}.
    R-50, E-B5, and Swin-L are short for ResNet-50, EfficientNet-B5~\citep{efficientnet} and Swin-Large, respectively.
    R-50-C$_{1}$ is the first block of the ResNet-50.
    $\dagger$ and $\ddagger$ indicate that the models are pre-trained by ImageNet-22K and additional depth estimation dataset, respectively.
    $^{*}$ indicates that the model is trained with extra class information.
    }
    \label{tab:nyu}
\end{table*}
\setlength{\tabcolsep}{6pt}
\setlength{\tabcolsep}{12pt}
\begin{table*}[t]
    \centering
    \resizebox{1.0\textwidth}{!}{
    \begin{tabular}{c|ccccccc}
        \toprule
        Method & Abs Rel$\downarrow$ & Sq Rel$\downarrow$ & RMSE$\downarrow$ & RMSE log$\downarrow$ & $\delta_{1}\uparrow$ & $\delta_{2}\uparrow$ & $\delta_{3}\uparrow$ \\
        \midrule
        DORN~\citep{dorn} & 0.072 & 0.307 & 2.727 & 0.120 & 0.932 & 0.984 & 0.994 \\
        BTS~\citep{bts} & 0.059 & 0.245 & 2.756 & 0.096 & 0.956 & 0.993 & \underline{0.998} \\
        TransDepth~\citep{TransDepth} & 0.064 & 0.252 & 2.755 & 0.098 & 0.956 & 0.994 & \textbf{0.999} \\ 
        DPT~\citep{vitdepth} & 0.062 & - & 2.573 & 0.092 & 0.959 & \underline{0.995} & \textbf{0.999} \\
        Adabins~\citep{adabins} & 0.058 & 0.190 & 2.360 & 0.088 & 0.964 & \underline{0.995} & \textbf{0.999} \\
        \midrule
        NeWCRFs~\citep{newcrf} & \underline{0.052} & 0.155 & 2.129 & \underline{0.079} & 0.974 & \textbf{0.997} & \textbf{0.999} \\ 
        DepthFormer~\citep{depthformer} & \underline{0.052} & 0.158 & 2.143 & \underline{0.079} & \underline{0.975} & \textbf{0.997} & \textbf{0.999} \\ 
        BinsFormer~\citep{binsformer} & \underline{0.052} & \underline{0.151} & \underline{2.098} & \underline{0.079} & 0.974 & \textbf{0.997} & \textbf{0.999} \\
        \midrule
        \textbf{RED-T (Ours)} & \textbf{0.050} & \textbf{0.146} & \textbf{2.080} & \textbf{0.077} & \textbf{0.976} & \textbf{0.997} & \textbf{0.999} \\
        \bottomrule
    \end{tabular}
    }
    \caption{Depth estimation performance on \textbf{KITTI Eigen split} dataset.
    }
    \label{tab:kitti_eigen}
    \vspace{-0.1cm}
\end{table*}
\setlength{\tabcolsep}{6pt}

\subsection{Dataset}\label{ssec:dataset} 

\vspace{0.1cm}
\textbf{NYU-v2}~\citep{nyu} dataset includes pairs of RGB images and depth maps on 464 indoor scenes, which are separated into 120K training samples from 249 scenes and 654 testing samples from 215 scenes. 
The range of depth labels is up to 10 meters.
We train our model on 50K subset following previous work~\citep{newcrf}.

\vspace{0.2cm}
\noindent\textbf{KITTI}~\citep{kitti} dataset consists of paired RGB images and corresponding depth maps obtained by a 3D laser scanner on 61 outdoor scenes while driving.
The range of depth annotations is up to 80 meters.
We apply two mainly used training/testing dataset splits. 
First, following the Eigen split setting~\citep{Eigen}, we train our model with about 26K samples from 32 scenes and test on 687 samples from 29 scenes.
Second, for the online depth prediction configuration~\citep{Kitti_split}, we use 72K training samples, 6K validation samples, and 500 testing samples.

\subsection{Implementation Details}\label{ssec:implementation}

We employ Swin-Large as a backbone, pre-trained on ImageNet-22K dataset~\citep{imagenet} with an input image size of 224 and window size of 7.
Each stage of the convolutional neck produces 512-channel feature maps, which are then concatenated and projected to 512 channels.
The number of depth-relative SA heads is set to 8 and their window size and shift size is set to 8 and 4, respectively.
We set $K=3$ and $N_b=128$ as default.
The size of the output depth map is the $1/4$ scale of the input image, which is then resized to the full resolution.

We use AdamW optimizer~\citep{adamw} with a learning rate of 1e-4, ($\beta_{1}$, $\beta_{2}$) of (0.9, 0.999), and a weight decay of 0.1.
The learning rate starts at 4e-6, increases to the maximum value for 25\% of the total iterations, and then decreases to 1e-6.
We train our model with a batch size of 16 for 24 epochs on $8\times$ NVIDIA A5000 24GB GPUs.
The gradient is accumulated every 2 batches and clipped to the maximum gradient norm of 0.1.
Please see the Appendix for details about data pre-processing, augmentation, metrics, and evaluation procedure.

\subsection{Depth Estimation Performance}\label{ssec:performance}

Table~\ref{tab:nyu} shows the MDE performance on the NYU-v2 dataset.
Despite the fact that several models employ the same or larger backbones than RED-T or exploit additional data during training~\citep{vitdepth}, RED-T achieves higher or comparable results in most of the metrics.
In particular, RED-T reduces `Abs Rel' and `log 10' errors by 4.2\% and 4.9\% compared to NeWCRFs~\citep{newcrf}, respectively.

Table~\ref{tab:kitti_eigen} presents the performance on KITTI Eigen split dataset.
RED-T outperforms previous works in every metric; especially, RED-T achieves lower relative errors (`Abs Rel' and `Sq Rel') and absolute errors (`RMSE' and `RMSE log').
We also evaluate RED-T on the KITTI official split which measures the performance on the official server.
As shown in Table~\ref{tab:kitti_online}, RED-T surpasses competitors by a large margin, especially in `Abs Rel' and `iRMSE' metrics.

The number of parameters of models that use the same Swin-L backbone is 270.4M, 273.8M, and 248.3M for NewCRFs, DepthFormer, and RED-T, respectively.
Note that the backbone contains 195.0M parameters.

\begin{table}[t]
    \centering
    \resizebox{1.0\linewidth}{!}{
    \begin{tabular}{c|cccc}
        \toprule
        Method & SILog $\downarrow$ & Abs Rel$\downarrow$ & Sq Rel $\downarrow$ &  iRMSE $\downarrow$ \\
        \midrule
        DORN & 11.77 & 8.78 & 2.23 &  12.98 \\
        BTS & 11.67 & 9.04 & 2.21 &  12.23 \\
        BANet & 11.55 & 9.34 & 2.31 &  12.17 \\
        PWA & 11.45 & 9.05 & 2.30 &  12.32 \\
        \midrule
        NeWCRFs & \underline{10.39} & \underline{8.37} & \underline{1.83} &  \underline{11.03} \\ 
        DepthFormer & 10.46 & 8.54 & \textbf{1.82} &  11.17 \\ 
        \midrule
        \textbf{RED-T (Ours)} & \textbf{10.36} & \textbf{8.11} & 1.92 & \textbf{10.82} \\
        \bottomrule
    \end{tabular}
    }
    \caption{Depth estimation performances on \textbf{KITTI official split}. 
    }
    \vspace{-0.1cm}
    \label{tab:kitti_online}
\end{table}

\subsection{Qualitative Evaluation}\label{ssec:qualitative}

In Figure~\ref{fig:comparison}(a), a truck is painted with diverse colors (i.e, black, gray, white) on its surface.
Although the depth of the surface continuously changes, in previous work~\citep{adabins}, undesired change in depth appears in the estimated output due to the color difference.
Another example is a kitchen counter wall decorated with a square pattern (Figure~\ref{fig:comparison}(c)).
While the depth of the wall should change smoothly, in the previous work, the pattern erroneously stands out in the depth map.
Thanks to the relative depth that help distinguish visual pits from visual hints, RED-T is robust to such obstacles. 
In other words, RED-T can accurately predict the depth of an object while much less affected by its visual appearance in 2D images.
Please check the Appendix for more qualitative comparisons.

\section{Range-restricted MDE}\label{sec:analysis}

\begin{figure}[h]
    \centering
    \begin{subfigure}[t]{0.485\linewidth}\centering
    \includegraphics[width=\linewidth]{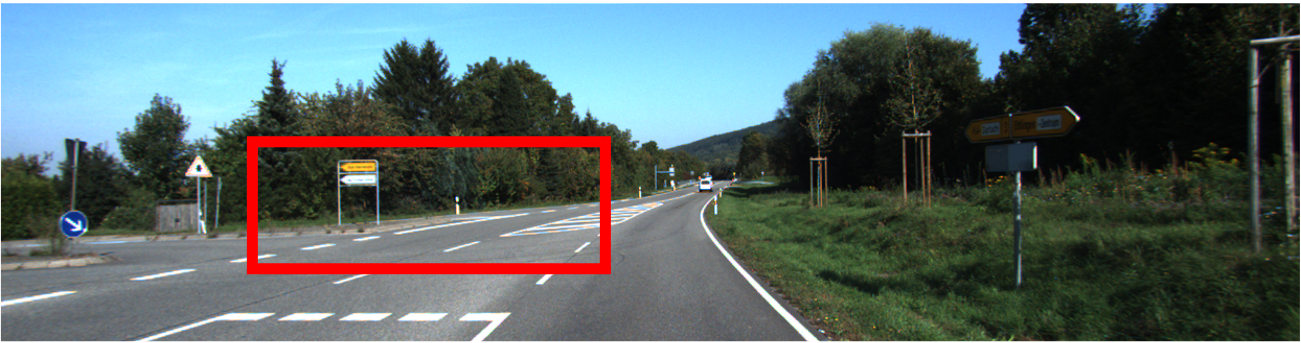}
    \caption{RGB Image}
    \vspace{0.1cm}
    \end{subfigure}
    \begin{subfigure}[t]{0.485\linewidth}\centering
    \includegraphics[width=\linewidth]{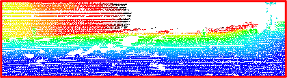}
    \caption{$d_{\text{clip}}= 80m$ (GT)}
    \vspace{0.1cm}
    \end{subfigure}
    \begin{subfigure}[t]{0.485\linewidth}\centering
    \includegraphics[width=\linewidth]{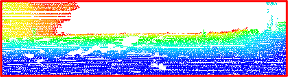}
    \caption{$d_{\text{clip}}= 60m$}
    \vspace{0.1cm}
    \end{subfigure}
    \begin{subfigure}[t]{0.485\linewidth}\centering
    \includegraphics[width=\linewidth]{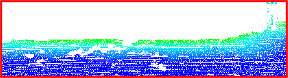}
    \caption{$d_{\text{clip}}= 40m$}
    \vspace{0.1cm}
    \end{subfigure}
    \caption{
    Examples of the range-restricted depth maps. GT implies ground truth, the actual label.
    In (c) and (d), labels larger than the threshold $d_{\text{clip}}$ are erased.
    All three correspond to the selected region (red box) in (a).
    }
    \label{fig:restricted_sample}
    \vspace{-0.1cm}
\end{figure}

\subsection{Motivation}\label{ssec:r_mde_motivation}

As mentioned in Section~\ref{sec:intro}, the harm of visual pits would be amplified when the model only exploits RGB information for depth estimation.
Unfortunately, this is an inherent problem for MDE because 1) the model only takes a single RGB image as input, and 2) the range of the annotated depth label is limited.
Therefore, for certain depth ranges that the model did not observe during training, the model solely depends on RGB values including visual pits which hurts the performance.

\subsection{Task Specification}\label{ssec:r_mde}

We propose a new MDE task that only a limited range of GT labels is given during training.
Specifically, let the GT labels in test data $d^{*} \in [d_{\text{min}} , d_{\text{max}}]$, then we remove labels larger than $d_{\text{clip}}$ and use only $[d_{\text{min}} , d_{\text{clip}}]$ during training phase.
As a result, the model should predict both seen and unseen depth ranges during the test phase.
Figure~\ref{fig:restricted_sample} shows examples of the restricted GT maps corresponding to different $d_{\text{clip}}$ values.

\begin{table}[t]
    \centering
    \resizebox{1.0\linewidth}{!}{
    \begin{tabular}{cc|llc}
        \toprule 
        Metric       & Model   & $\sim$40\textit{m}  & $\sim$60\textit{m}   & $\sim$80\textit{m}   \\
        \midrule
        \multirow{3}{*}{Abs Rel$\downarrow$ }
                    & AdaBins                   & 0.091 (+56.9\%)     & 0.077 (+32.8\%)    & 0.058     \\
                    & NeWCRFs                   & 0.058 (+11.5\%)     &  0.054 (+3.8\%)    & 0.052     \\
                    & \textbf{RED-T}     &   \textbf{0.055 (+10.0\%)}   & \textbf{0.050 (+0.0\%)}     & \textbf{0.050}     \\
        \midrule
        \multirow{3}{*}{RMSE$\downarrow$}
                    & AdaBins                   & 4.048(+70.4\%)     & 2.697 (+13.6\%)     & 2.375     \\
                    & NeWCRFs                   & 3.616 (+69.8\%)    & 2.336 (+9.7\%)      & 2.129     \\
                    & \textbf{RED-T}     & \textbf{3.212 (+54.4\%)}     & \textbf{2.232 (+7.3\%)}     & \textbf{2.080}     \\
        \midrule
        \multirow{3}{*}{$\delta_{1}\uparrow$ }
                    & AdaBins                   & 0.935 (-3.0\%)     & 0.956 (-0.8\%)    & 0.964     \\
                    & NeWCRFs                   & 0.955 (-2.0\%)     & 0.970  (-0.4\%)   & 0.974     \\
                    & \textbf{RED-T}    & \textbf{0.957 (-1.9\%)}     & \textbf{0.974 (-0.2\%)}    & \textbf{0.976}     \\
        \bottomrule
    \end{tabular}
    }
    \caption{Depth estimation performance on KITTI dataset using only a limited range of depth labels.
    The last three columns represent the maximum observable depth value ($d_{\text{clip}}$) during training.
    }
    \vspace{0.1cm}
    \label{tab:restricted}
\end{table}

\setlength{\tabcolsep}{11pt}
\begin{table}[t]
    \centering
    \resizebox{1.0\linewidth}{!}{
    \begin{tabular}{cc|llll}
        \toprule 
        Metric & Rel.bias & $\sim$2\textit{m}  & $\sim$4\textit{m}  & $\sim$6\textit{m}  & $\sim$8\textit{m}  \\
        \midrule
        \multirow{2}{*}{Abs Rel$\downarrow$ }
                    & \xmark & 0.365  & 0.109   & 0.094     & 0.091      \\
                    & \cmark & 0.307  & 0.106   & 0.092     & 0.091      \\
        \midrule
        \multirow{2}{*}{RMSE$\downarrow$}
                    & \xmark & 1.882  & 0.533   & 0.366     & 0.337      \\
                    & \cmark & 1.537  & 0.504   & 0.360     & 0.331      \\
        \midrule
        \multirow{2}{*}{$\delta_{1}\uparrow$ }
                    & \xmark & 0.471  & 0.868   & 0.917     & 0.925      \\
                    & \cmark & 0.491  & 0.878   & 0.919     & 0.925      \\
        \bottomrule
    \end{tabular}
    }
    \caption{MDE performance on NYU-v2 dataset using only a limited range of depth labels.
    A variant of RED-T without depth-relative attention bias (Rel.bias) is compared.
    }
    \vspace{-0.1cm}
    \label{tab:restricted_nyu}
\end{table}
\setlength{\tabcolsep}{6pt}

\begin{figure}[ht]
    \centering
    \begin{subfigure}[t]{0.850\linewidth}\centering  
    \includegraphics[width=\linewidth]{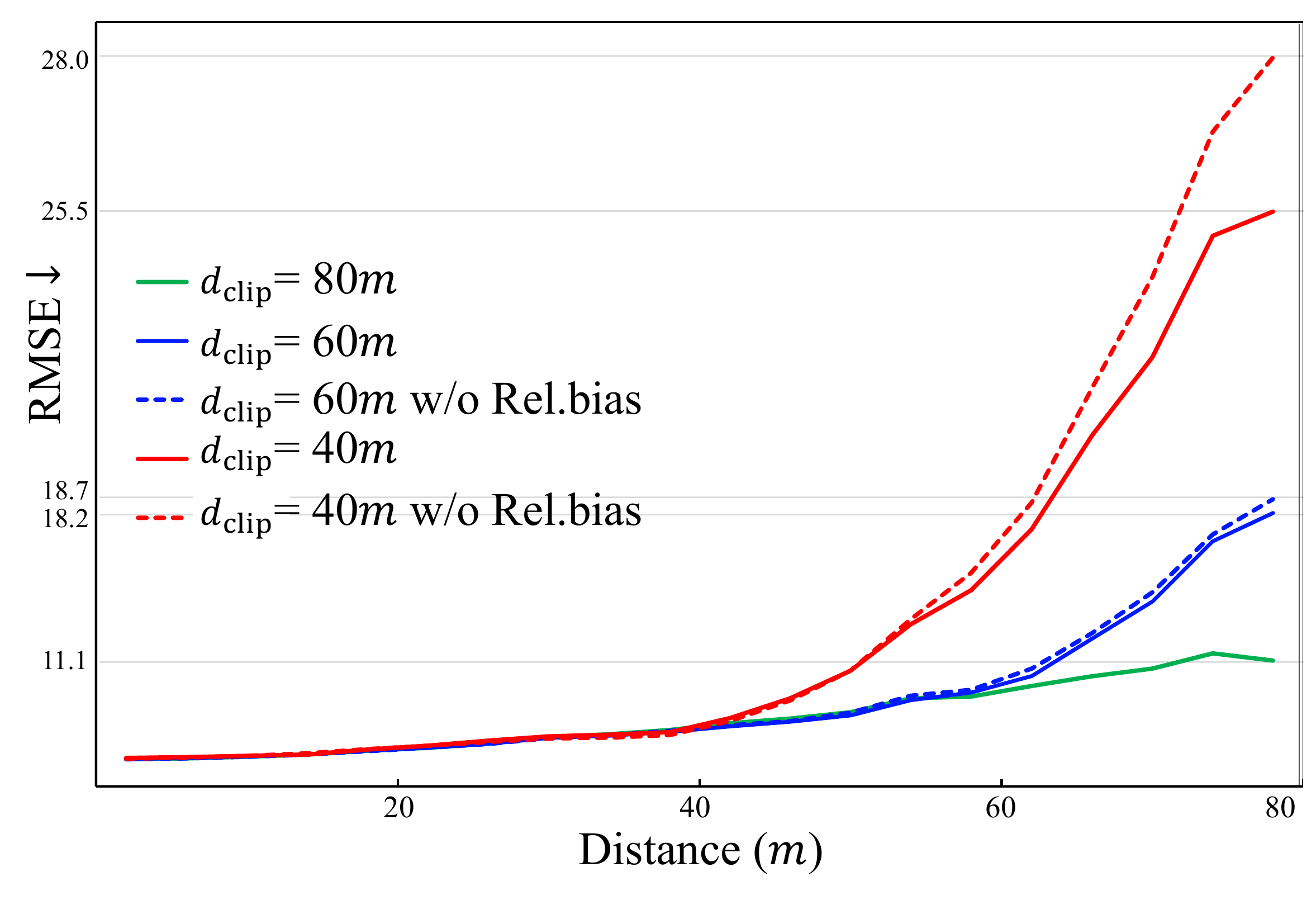}
    \caption{Comparison between models trained with different $d_{\text{clip}}$ (solid) and models without the relative bias (dashed).}
    \vspace{0.1cm}
    \end{subfigure}
    \begin{subfigure}[t]{0.860\linewidth}\centering 
    \includegraphics[width=\linewidth]{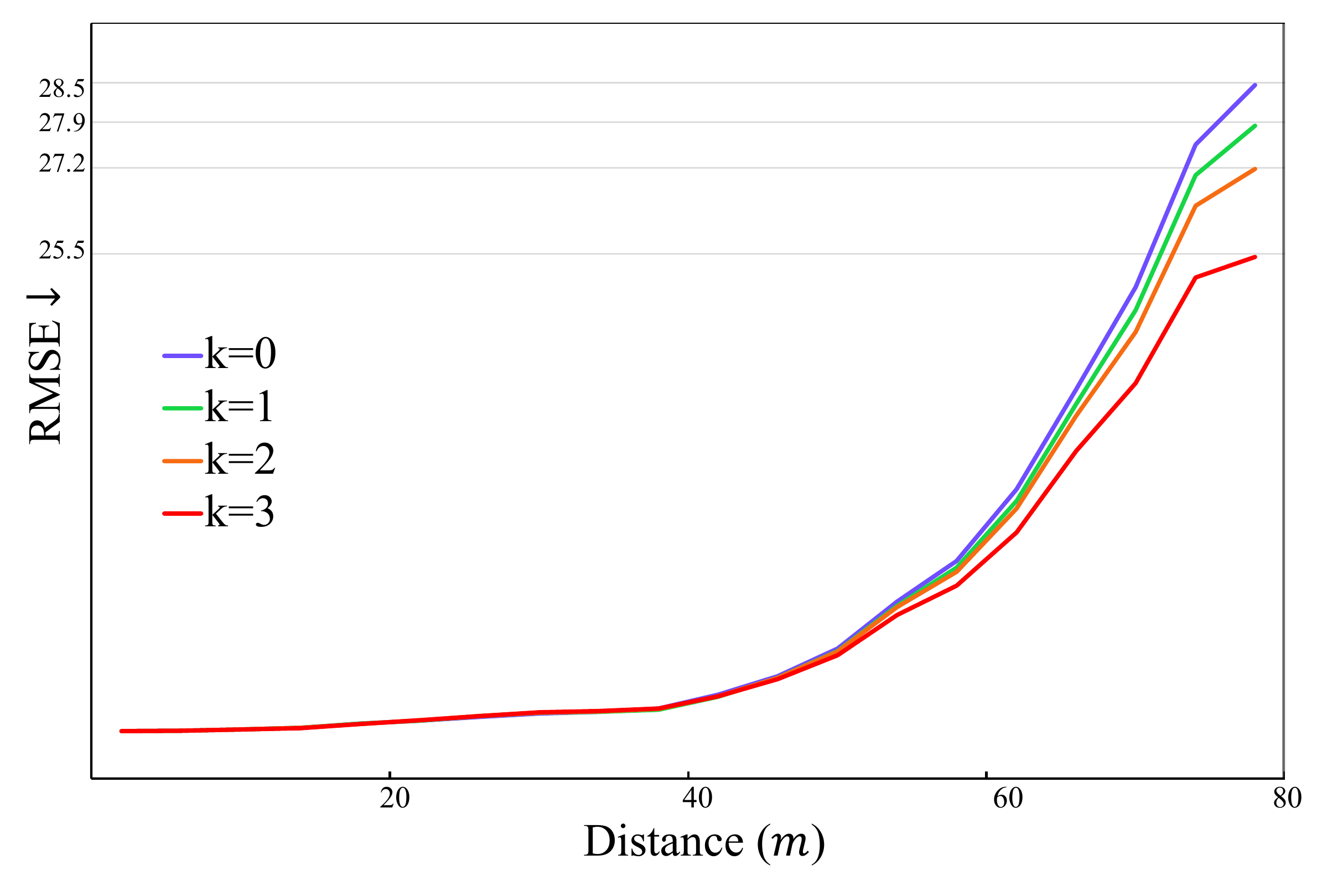}
    \caption{Comparison between the depth estimation performance of intermediate depth maps $D_k$ in $d_\text{clip}$=40m setting.}
    \vspace{0.1cm}
    \end{subfigure}
    \caption{
    Fine-grained depth estimation results of range-restricted MDE experiments on the KITTI dataset.
    }
    \vspace{-0.4cm}
    \label{fig:per_range}
\end{figure}

\subsection{Experimental Results}\label{ssec:range}

We conduct experiments on the KITTI dataset, where $d_{\text{max}}=80m$, with two configurations of $d_{\text{clip}} \in \{ 40m, 60m \}$.
Table~\ref{tab:restricted} shows that RED-T achieves much lower performance degradation than previous models.
In $d_{\text{clip}}=60m$ setting, RED-T achieves zero performance loss in the `Abs Rel' metric and 7.3\% reduction in `RMSE' metric, while AdaBins suffers from 32.8\% and 13.6\% performance loss, respectively.
We claim that RED-T is robust to unseen depth range because the model can avoid visual pits by actively incorporating the relative depth information in the model design.

\subsection{Effectiveness of Relative Bias}\label{ssec:feature}

To highlight the importance of the relative depth, we repeat the same experiments without relative bias (i.e., forcing $R=0$) on the NYU dataset.
In Table~\ref{tab:restricted_nyu}, RED-T without depth-relative attention guidance shows 1) worse performance and 2) larger relative performance decay compared to the proposed RED-T.
The gap between the RED-T with and without relative bias becomes larger as the observable depth range ($d_{\text{clip}}$) decreases.

In addition, we measure the RMSE as a function of distance in various scenarios.
In Figure~\ref{fig:per_range}(a), models trained by a restricted depth range show much larger error compared to the baseline (i.e, model trained on full depth range) in unseen (depth) ranges.
We observe that relative bias improves the generalization capability of the model in unseen ranges.
Furthermore, in Figure~\ref{fig:per_range}(b), we show that multiple iterations of depth-relative processing in the head consistently reduce the error.
Specifically, the RMSE is reduced by 2.1\%, 4.6\%, and 10.5\% as the number of iterations $k$ increases.
 In fact, even for the $K$=1 case (without multiple refine stages), the proposed RED-T outperforms competitors.

One may think that the improvement of the relative bias is not dramatic on conventional depth estimation metrics.
We argue that current metrics do not sufficiently express the negative effect of visual pits.
First, the metric values are averaged over valid pixels that are sparsely annotated, but visual pits mostly appear within concentrated regions.
Second, in terms of the number of pixels, the proportion of visual pits to the entire image is often very small (under 1\% over the entire image). 
Nevertheless, we emphasize that visual pits are risk factors for practical systems; even the danger amplifies when the model attempts to predict unseen depth.
\section{Conclusion}\label{sec:conclusion}

In this paper, we proposed RED-T which aims at minimizing the adversarial effect of visual pits.
To do so, RED-T utilizes relative depth information as a means to guide the monocular depth estimation process.
Specifically, we adopt self-attention bias to encourage each pixel to assign high attention weight to other pixels of close depth.
RED-T achieved superior depth estimation performance on NYU-v2, KITTI Eigen/official split datasets compared to the competitors.
To demonstrate the effectiveness of relative depth, we introduced a new MDE task that restricts observable depth range during training.

\section*{Acknowledgments}
This work was supported by the Future-promising Convergence Technology Pioneer Program of the National Research Foundation of Korea (NRF) grant funded by the Korea Ministry of Science and ICT (MSIT) (No. 2022M3C1A3098746) and National Research Foundation of Korea (NRF) grant funded by the Korea government (MSIT) (RS-2023-00208985).

\newpage

\bibliographystyle{named}
\bibliography{ijcai23}

\newpage  
\appendix

\section{Details}\label{supp:sec:details}

\subsection{Data Processing}

The input image resolutions are (480, 640) and (352, 1216) for NYU-v2 and KITTI datasets, respectively.
For training data augmentation, we apply random rotation, left-right flip, brightness change, and color adjustment for both datasets, following common practice~\citep{adabins,newcrf}.
For the KITTI dataset, we apply a common region cropping (a.k.a. KB-crop) following previous works~\citep{adabins,newcrf,depthformer}, and randomly crop the image to 352$\times$704 during training.
For the NYU-v2 dataset, we do not apply random crop because we find that crop-and-resize affects the data statistics and leads to unstable training when used with the Swin backbone.
Please check the provided code for more details and settings.

\subsection{Evaluation and Metrics}

\vspace{0.1cm}\paragraph{Evaluation details}
For the NYU-v2 dataset, we evaluate models on the pre-defined center cropping introduced by Eigen~\citep{Eigen}.
For the KITTI Eigen split dataset, we use the crop suggested by Garg~\citep{Garg}.
We do not exploit the test time augmentation that combines the left-right flip results.

\paragraph{Metrics} 
We follow the standard evaluation protocol as prior works~\citep{dorn,bts}.
Table~\ref{tab:metric} summarizes the metrics for MDE.
We evaluate our model with the accuracy under threshold ($\delta_{i} < 1.25^{i}, i=1,2,3$), mean absolute relative error (Abs Rel), root mean squared error (RMSE), and mean log10 error (log 10) for NYU-v2 dataset.
We additionally measure root mean squared log error (Sq Rel) and root mean squared log error (RMSE log) for KITTI Eigen split dataset.

\subsection{Block Architectures}

\begin{figure*}[t]
    \centering
    \begin{subfigure}[t]{0.80\textwidth}\centering
        \includegraphics[width=\textwidth]{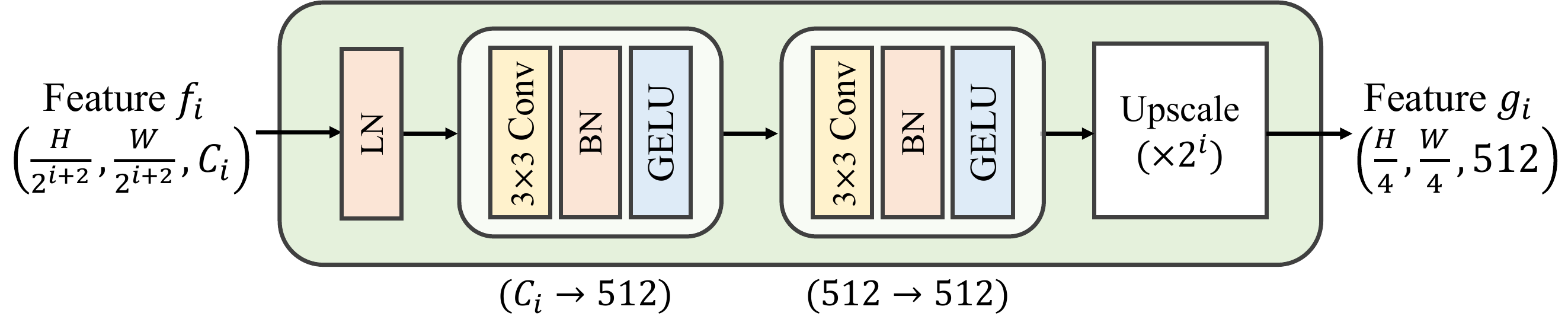}
        \caption{Convolutional Neck Block (CNB)}
        \vspace{0.4cm}
    \end{subfigure}
    \begin{subfigure}[t]{0.35\textwidth}\centering
        \includegraphics[width=\textwidth]{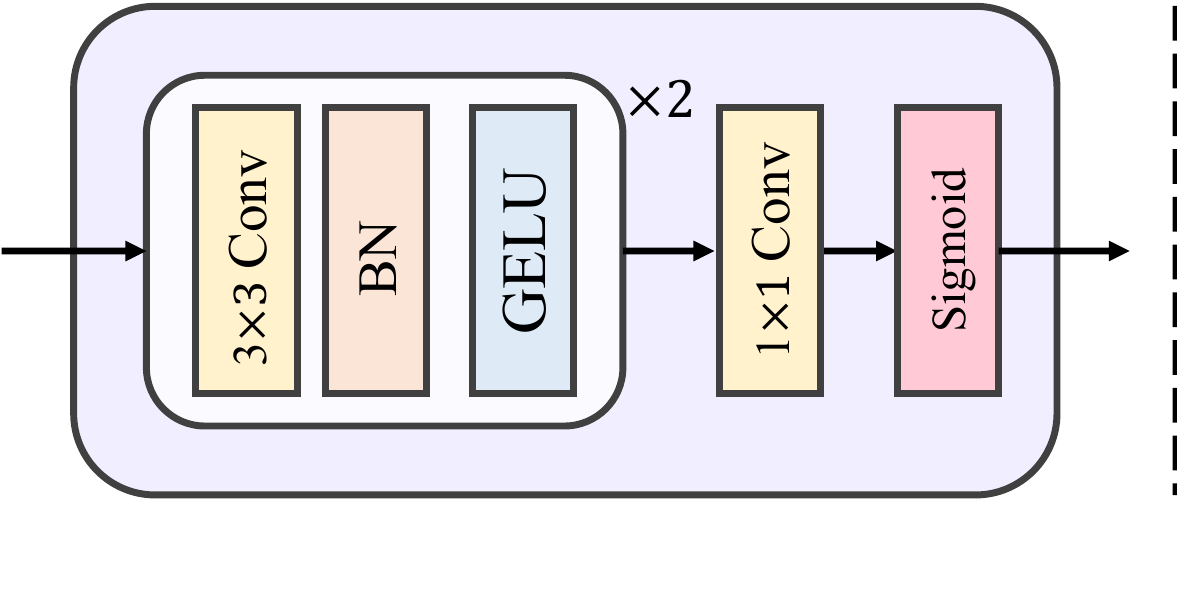}
        \caption{Depth Estimation Block (DEB)}
    \end{subfigure}
    \begin{subfigure}[t]{0.55\textwidth}\centering
        \includegraphics[width=\textwidth]{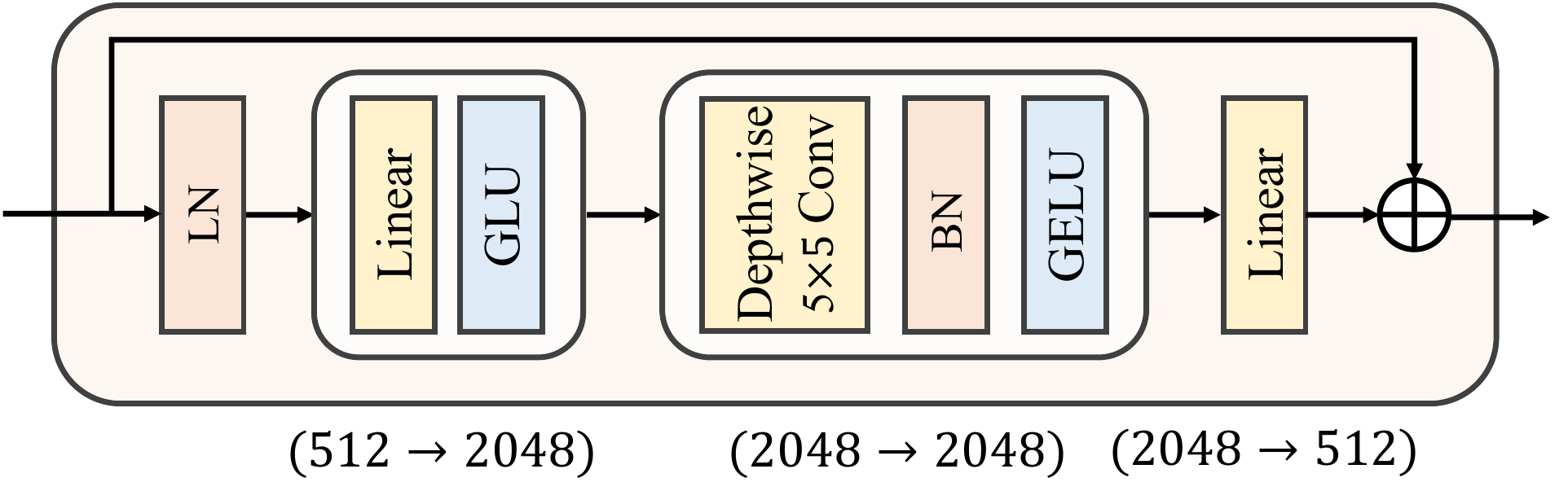}
        \caption{Convolutional Feedforward Block (CFF)}
    \end{subfigure}
    \caption{
    Illustration of (a) CNB, (b) DEB, and (c) CFF in Figure 2.
    BN and LN represent batch normalization and layer normalization layers, respectively.
    GLU indicates a gated linear unit~\citep{glu}.
    In (c), the residual connection adds the input and output.
    }
    \label{fig:additional_block}
    \vspace{0.1cm}
\end{figure*}

Figure~\ref{fig:additional_block} shows the details of each component explained in Section 2 (used in Figure 2).
Three components are illustrated: convolutional neck block (CNB), depth estimation block (DEB), and convolutional feedforward block (CFF).

\vspace{0.1cm}\paragraph{Convolutional Neck Block (CNB)} 
The output $\text{SA}_h$, a weighted sum of value vectors in $V_h$, does not contain positional information because a positional embedding is only used in attention weights $A_h$.
For this reason, the local dependency of Swin-based features can be much weaker than other CNN-based features~\citep{vit_analysis,vit_cnn,vit_vs_cnn}.
To enhance the locality of backbone features, we utilize convolutional layers for each feature in the neck.
Due to the parallel design, the model can first relate nearby pixels within each scale before integrating every multi-scale feature.

\vspace{0.1cm}\paragraph{Convolutional Feed-forward (CFF)}
Similar to the position-relative SA in Swin, depth-relative SA in the head also does not embed positional information into the output features.
Inspired by Conformer~\citep{conformer}, we alleviate this issue by inserting a depth-wise convolution layer in the middle of the feed-forward (FF) module.
By providing additional locality to the high-resolution features, convolution operations can improve the output depth map to be more consistent through neighboring pixels.

\vspace{0.1cm}\paragraph{Depth Estimation Block (DEB)}
Depth estimation block simply takes features and passes through three convolution layers. The output is scaled to [0, 1] range by the sigmoid function.

\section{Comparison}\label{supp:sec:qualitative}

\setlength{\tabcolsep}{15pt}
\begin{table}[t]
    \centering
    \renewcommand{\arraystretch}{1.6}
    \small
    \begin{tabular}{cc}
        \toprule
        Metric & Formulation \\
        \midrule
        Abs Rel & $\frac{1}{n}\sum\nolimits_{p}^n\frac{|D_{p}-D^{*}_{p}|}{D}$ \\
        RMSE & $\sqrt{\frac{1}{n}\sum\nolimits_{p}^n(D_{p}-D^{*}_{p})^2}$  \\
        RMSE log & $\sqrt{\frac{1}{n}\sum\nolimits_{p}^n\|d_{p}\|^2}$ \\
        log10 & $\frac{1}{n}\sum\nolimits_{p}^n|\log_{10} D_{p} - \log_{10} (D^{*}_{p})|$ \\
        Sq Rel & $\frac{1}{n}\sum\nolimits_{p}^{n}\frac{\|D_{p}-D^{*}_{p}\|^{2}}{D} $\\ 
        SILog & $\frac{1}{n}\sum\nolimits_{p}^n d^{2}_{p} - \frac{1}{n^2}(\sum\nolimits_{p}^n d_{p})^2$\\
        \bottomrule
    \end{tabular}
    \caption{Formulations of depth estimation metrics.
    $d_{p}=\log{D_{p}} - \log{D^{*}_{p}}$ denotes the logarithmic depth difference between the predicted depth and the ground truth depth.
    }
    \label{tab:metric}
    \vspace{-0.2cm}
\end{table}
\setlength{\tabcolsep}{6pt}  

\subsection{Qualitative Comparison}

\begin{figure*}[t]
    \centering
    \begin{subfigure}{0.71\textwidth}
        \includegraphics[width=1.0\linewidth]{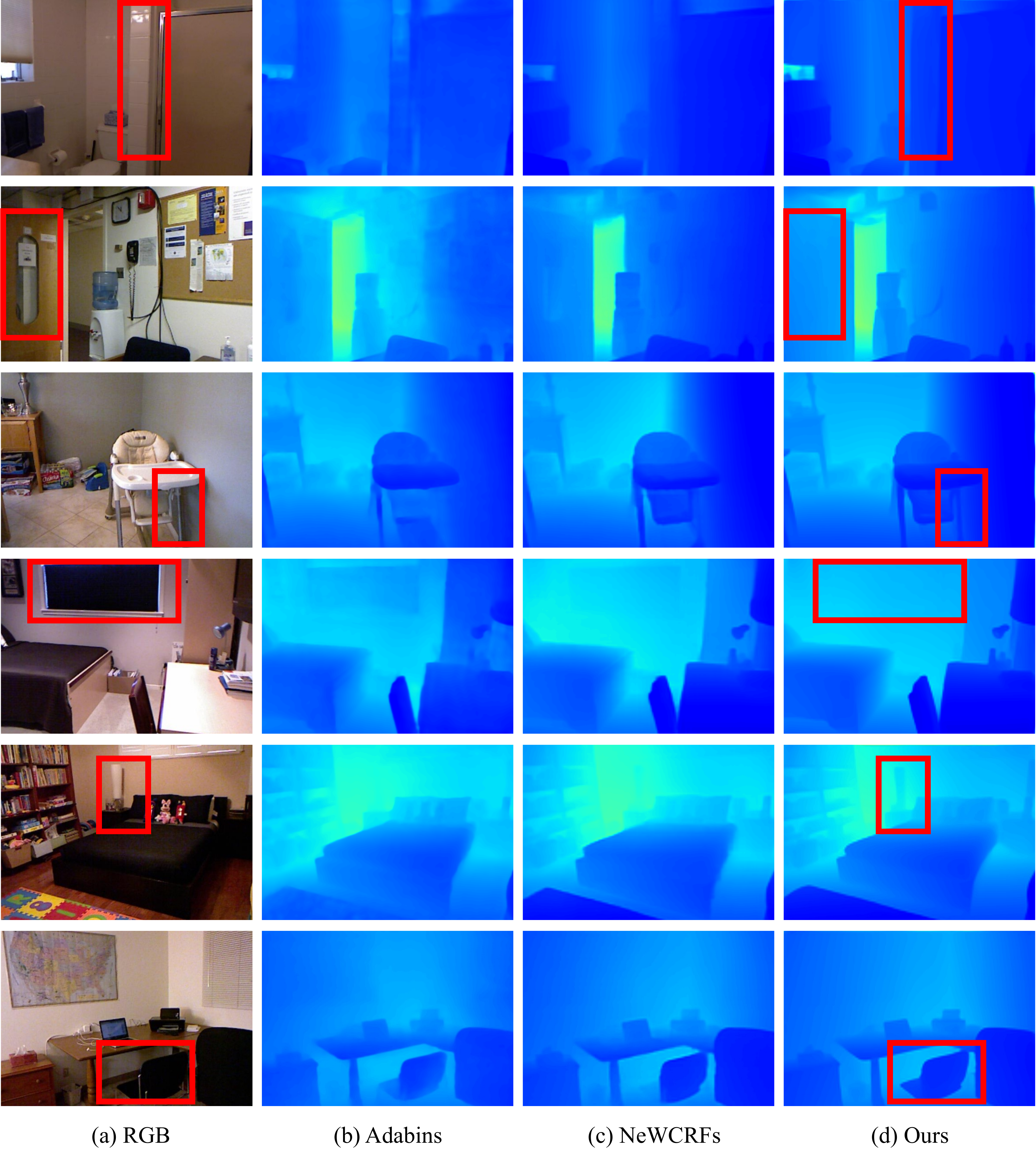}
        \caption{Qualitative comparison on the NYU-v2 dataset.}
        \vspace{0.5cm}
        \label{fig:qual_nyu}
    \end{subfigure}
    \begin{subfigure}{0.71\textwidth}
        \includegraphics[width=1.0\linewidth]{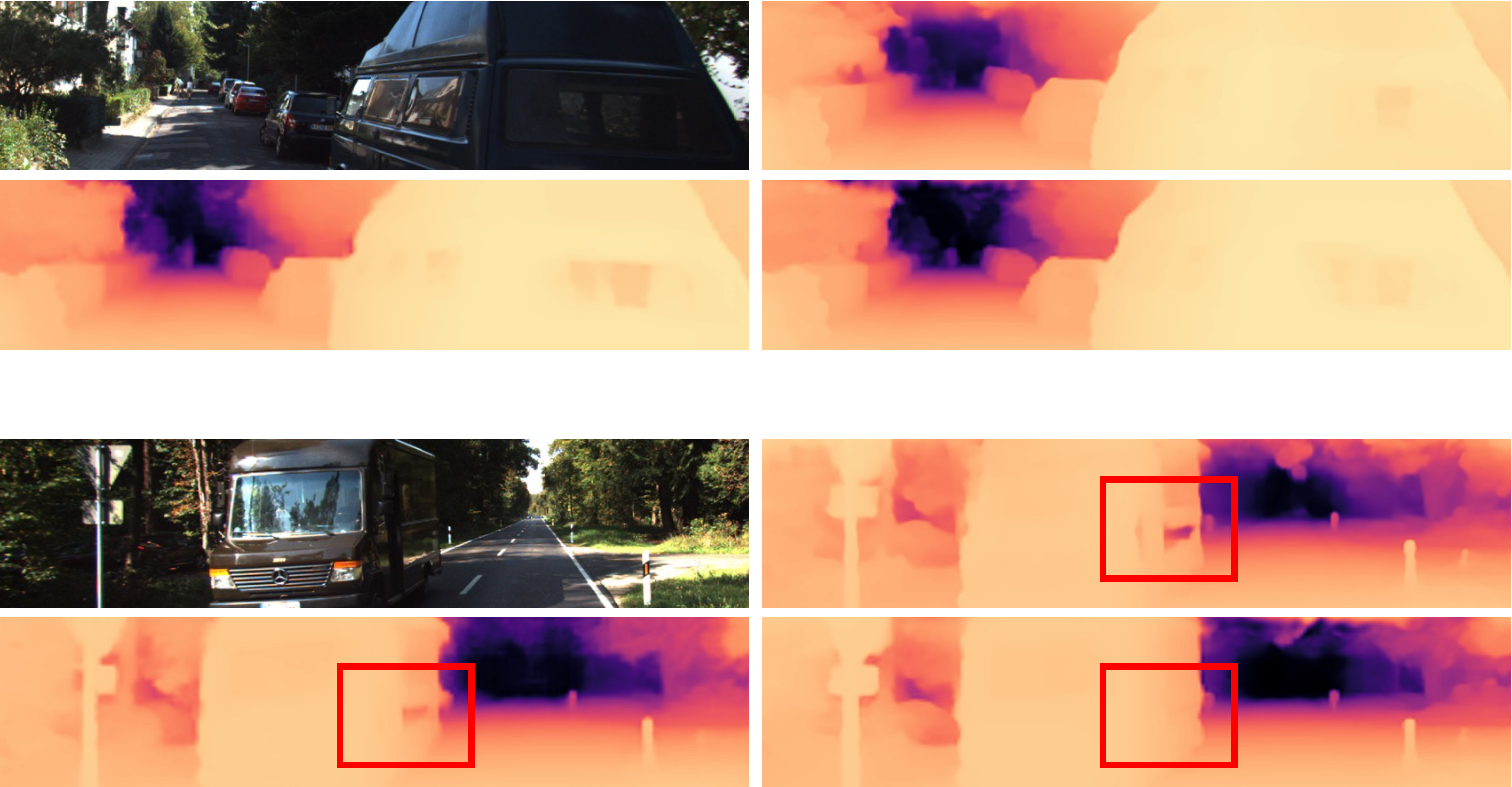}
        \caption{Qualitative comparison on the KITTI dataset. 
        The order of four images in a clockwise direction is as follows: RGB, AdaBins, Ours, and NeWCRFs.}
        \label{fig:qual_kitti}
    \end{subfigure}
    \label{fig:qual}
\end{figure*}

We present additional qualitative results in Figure~\ref{fig:qual_nyu} (NYU-v2) and Figure~\ref{fig:qual_kitti} (KITTI).
Specifically, we compare the results from the proposed RED-T with previous state-of-the-art models, AdaBins~\citep{adabins} and NewCRFs~\citep{newcrf}.
The comparison demonstrates that previous works are highly affected by visual pits such as reflection, mirror, shadow, and color.
In contrast, RED-T shows consistent depth prediction on a flat surface, regardless of its visually diverse appearance.
Furthermore, we note that the object boundary of the predicted depth map is sharper and better aligned with the edge of the actual object compared to previous works.


\subsection{Computation Comparison}

We also compare three MDE architectures in terms of inference speed (Throughput) and resource usage (\#Parameters).
The metrics are measured on a single NVIDIA RTX-3070 GPU with a batch size of 1 and KITTI image size of (352, 1216).
\setlength{\tabcolsep}{7pt}
\begin{table}[t]
    \centering
    \renewcommand{\arraystretch}{1.2}
    \vspace{-0.1cm}
    \resizebox{1.0\linewidth}{!}{
    \begin{tabular}{c|ccc}
    \toprule
        Metric & Adabins & NeWCRFs & RED-T (ours) \\
        \midrule
        \#Params (M) & 78 & 270 & 248 \\
        Throughput (img/sec) & 17 & 14 & 10 \\
    \bottomrule
    \end{tabular}}
    \vspace{-0.1cm}
    \caption{Computational comparison of MDE models.}
    \label{tab:comparison}
    \vspace{-0.1cm}
\end{table}
\setlength{\tabcolsep}{6pt}

As expected, AdaBins exhibits the highest throughput (i.e., the shortest inference time) due to its relatively lightweight backbone.
Unfortunately, even the smallest AdaBins model should use GPU-based systems to achieve real-time inference speed (e.g., 20-30 frames per sec).
This means that recently proposed MDE models may not be feasible for CPU-only inference in performance-critical areas such as autonomous driving.

Furthermore, we would like to highlight the potential benefits of the large backbone: multitasking capability.
In autonomous driving, for example, multiple vision tasks should be performed simultaneously.
In this case, an approach to run a large backbone only once and then exploit rich features generated from it for various task-specific heads can save a considerable amount of `redundant' computations.

\section{Ablation}\label{supp:sec:ablation}

We first emphasize that Table~\ref{tab:restricted_nyu} and Figure~\ref{fig:per_range} clearly demonstrate the effectiveness of depth-relative bias in RED-T, as stated in Section 4.4.
We have conducted additional ablation studies on the architectural components (KITTI Eigen split, full 80m range).
The results indicate that our depth-relative bias indeed contributes to the performance.
Additionally, we observe that our neck and head designs outperform the conventional feature pyramid network (FPN) and simple projection-regression layer.
\setlength{\tabcolsep}{14pt}
\begin{table}[t]
    \centering
    \renewcommand{\arraystretch}{1.2}
    \resizebox{1.0\linewidth}{!}{
    \begin{tabular}{ccc|cc}
        \toprule
        Neck & Head & Rel.bias & Sq Rel$\downarrow$ & RMSE$\downarrow$ \\
        \midrule
        FPN & Proj. & n/a & 0.151 & 2.128 \\
        FPN & \textbf{Ours} & \cmark & 0.150 & 2.102 \\
        \midrule
        \textbf{Ours} & Proj. & n/a & 0.150 & 2.112 \\
        \textbf{Ours} & \textbf{Ours} & \xmark & 0.148 & 2.085 \\
        \textbf{Ours} & \textbf{Ours} & \cmark & \textbf{0.146} & \textbf{2.080} \\
        \bottomrule
    \end{tabular}}
    \caption{Ablation on different neck and head architectures (KITTI).}
    \vspace{-0.3cm}
    \label{tab:ablation_component}
\end{table}
\setlength{\tabcolsep}{6pt}

\section{Discussion}\label{supp:sec:discussion}

\subsection{Relative Depth}

Although the proposed depth-relative attention exploits the term \textit{relative depth}, our model estimates the per-pixel absolute (ordinary) depths, not relatively scaled depths.
Specifically, the latter is often known as the relative depth estimation task~\citep{rel_depth_est}.
We follow the previous works~\citep{depth_cues,rel_depth_aug} that also use the term relative depth as a difference between depths.

\subsection{Sparse Depth Label}
In many depth estimation datasets, ground truth (GT) depth labels are sparsely annotated due to the hardware limitation of depth sensors, such as LiDAR, Radar, Structured-Light, and Time-of-Flight~\citep{ttf,sparse_guide_s3,pseudo_lidar}.
To better utilize this sparse information, approaches to incorporate sparse GT as additional input have been proposed~\citep{sparse_pixel,sparse_supervision,sparse_to_dense}.
This task, also known as depth completion, differs from MDE because an additional sparse depth map is used as an input to supplement RGB information.
In our proposed MDE setup, we also sparsify the depth map but it is only used as a label (not input) in the training process.
Our sparsification strategy is also distinct from previous works that uniformly remove pixels through the entire depth range~\citep{sparse_pixel} since we sparsify the label by restricting the observable depth range during the training.
When compared to previously studied sparse setups, the proposed setup is far more challenging because the distribution of GT labels significantly differs for training and test phases.

\subsection{Depth Densification and Synthetic Label}
In order to reduce the negative effect of sparse labels (i.e., LiDAR), previous studies have exploited external sources of data.
One can incorporate synthetic images into the training process to make the MDE model more robust for visual pits and unseen range~\citep{zheng2018t2net,zhao2019geometry,cheng2020s,pnvr2020sharingan,gurram2021monocular}.
Synthetic images produce fully annotated dense depth labels without error and can generate more informative scenarios and environments for training.
However, using only synthetic data cannot improve the result because of the input domain gap; in other words, the models trained with synthetic images often require a domain adaptation process to mitigate the domain mismatch problem.
The abovementioned works combine real and synthetic data and achieve considerable improvement compared to using only sparse-and-real data.
In particular, MonoDEVSNet~\citep{gurram2021monocular} integrates synthetic data (i.e., virtual world) supervision and Structure-from-Motion (SfM) (i.e., real-world) self-supervision, outperforming AdaBins by a considerable margin.

\subsection{DNNs Suffer from Visual Pits}
In many cases, DNNs suffers from learning unwanted biased representations. 
For example, CNN-based models tend to be biased toward texture than shape~\citep{bias_imagenet,bias_cnn}, while ViT exhibits the opposite behavior~\citep{bias_vit}.
Both CNN and ViT are still too sensitive to dataset-dependent characteristics~\citep{spurious_vit,spurious_cnn}.
Our paper (and introducing the concept of \textit{visual pit}) can reduce the influence of unnecessary information and thus improve model robustness to such pitfalls.
Please note that we are the first to point out and address this problem in the context of depth estimation literature.

\subsection{Limitation and Ethics Statement}

While RED-T achieves state-of-the-art performance on MDE tasks, the model has potential limitations.
First, Transformer-based models take longer training and inference time than CNN-based ones.
Second, a powerful backbone extracts rich and diverse visual hints but also increases the potentially harmful visual pits.
We utilized relative depth information in the head to reduce the negative effect of visual pits, however, we could not filter out visual pits from the backbone.
Third, during the iterative depth map refinement process, depth discretization and embedding indexing are non-differentiable operations, making each intermediate depth map a leaf node of the computational flow.
We expect further performance improvement by approximating these operations to be differentiable.

We do not expect any ethical concerns for this paper.
On the other hand, we believe the proposed method can reduce the potential risk of autonomous driving systems.

\subsection{Future Work}

The proposed depth-relative attention mechanism can be applied to various depth estimation tasks, including self-supervised depth estimation and depth completion.
We believe the adverse influence of visual pits would also appear in other tasks, and the proposed method can mitigate the problem.
In addition, the relative depth information can be more actively utilized by designing a specialized loss function that directly employs the relative depth as the target.

\end{document}